\documentclass{article} 
\usepackage{mimir_arxiv,times}

\usepackage{amsmath,amsfonts,bm}

\def\eqref#1{equation~\ref{#1}}

\def\1{\bm{1}}

\DeclareMathAlphabet{\mathsfit}{\encodingdefault}{\sfdefault}{m}{sl}
\SetMathAlphabet{\mathsfit}{bold}{\encodingdefault}{\sfdefault}{bx}{n}

\usepackage{graphicx}
\usepackage{caption}
\usepackage{subcaption}
\usepackage{xspace}
\usepackage{enumitem}
\usepackage{multirow}
\usepackage{hyperref}
\usepackage{listings}
\usepackage{booktabs} 
\usepackage{tabularx}

\usepackage{url}            
\usepackage{amsfonts}       
\usepackage{nicefrac}       
\usepackage{microtype}      
\usepackage{xcolor}         
\usepackage{amsmath}
\usepackage{subcaption}
\usepackage{algorithm}
\usepackage{algorithmic}

\usepackage{wrapfig}
\newcommand{\ours}{\textsf{M\'{\i}mir}\xspace}

\title{\ours: Physics-Grounded LLM Agents for Long-Horizon Irrigation Control}

\author{
Yimeng Liu$^{1}$ \quad
Mi Zhang$^{2}$ \quad
Younsuk Dong$^{1}$ \quad
Zhichao Cao$^{1}$
\\[1.2ex]
$^{1}$ Michigan State University
\quad
$^{2}$ Ohio State University
}

\begin{document}

\maketitle
\begin{abstract}
Large language model (LLM) agents increasingly combine reasoning, tool use, and action, but most evidence comes from episodic tasks with relatively immediate feedback and reset failures. 
Long-running physical control operates in a different regime: actions alter future states, errors compound across decisions, and an agent must improve from experience without being allowed to rewrite the physical rules that make execution safe. 
We study this regime through irrigation, where daily decisions interact with soil-water dynamics over entire growing seasons. 
We present \ours, a physics-grounded LLM agent organized around two repair timescales. 
At the fast timescale, a structured physical interface and deterministic simulator turn an LLM output into a proposal that we numerically check, revise, and subject to bounded deterministic action selection before execution. 
At the slow timescale, recurrent failure patterns are consolidated into persistent contextual principles that condition future proposals, while the physical model, evaluator, and execution constraints remain immutable. 
Under a common retrospective evaluator across multiple sites, crops, and years, \ours attains the lowest reported aggregate control cost among the evaluated references and uses about 51\% less irrigation than the historical schedule replay. 
The ablation study show higher control cost when forward simulation, verified revision, or persistent context is removed; model-scale and model-family studies show no monotonic gain from increasing LLM size. 
The resulting lesson show that persistent physical agents can combine semantic reasoning with bounded, evidence-driven self-improvement while reserving physical truth and actuator authority for explicit numerical mechanisms.
\end{abstract}

\section{Introduction}
\label{sec:intro}

Large language models (LLMs) are increasingly used as agents that interleave reasoning, tool use, memory, and action~\cite{Yao2023ReAct}. 
In embodied settings, language models can ground high-level decisions in perception, affordances, and external tools~\cite{Ahn2022SayCan,Driess2023PalmE,huang2023innermonologue,huang2023voxposer}.
However, most agent evaluations remain episodic: a task starts, the agent acts for a finite horizon, and the episode terminates. 
Persistent physical control is different. 
Each action changes the state on which later decisions depend; consequences can be delayed; and a locally plausible mistake can persist through the physical system long after the language-model call that produced it. 
A physical agent must therefore do more than generate a reasonable action. 
It need to remain grounded in numerical dynamics, repair bad decisions before actuation, and accumulate useful experience without allowing that experience to erode execution safety.

Irrigation provides a concrete instance of this regime. 
A controller repeatedly decides how much water to apply under evolving soil, crop, weather, and forecast conditions. 
Irrigation changes root-zone water storage and therefore the context for future decisions; excessive application consumes water and can increase drainage, whereas insufficient application can accumulate crop-water deficit~\cite{Allen1998FAO56}. 
The effect of today's decision may only become evident after subsequent evapotranspiration, rainfall, and additional control actions. 
Physics-based scheduling, model predictive control (MPC), sensor-driven irrigation, and reinforcement learning already provide strong solutions when the state, model, objective, and controller are sufficiently specified~\cite{Allen1998FAO56,Delgoda2016IrrigationMPC,Camacho2013MPC,dong2024implementation,Yu2026IrrigationAgents,Ding2024DRLIC}. 
We therefore do \emph{not} ask whether an LLM should replace these numerical controllers. 
We ask a narrower systems-and-learning question: \emph{if an LLM is introduced as a reasoning component of a long-running physical controller, which responsibilities can be adaptive and which must remain explicit, numerical, and externally enforceable?}

\begin{wrapfigure}{r}{0.48\linewidth}
  \centering
  \includegraphics[width=\linewidth]{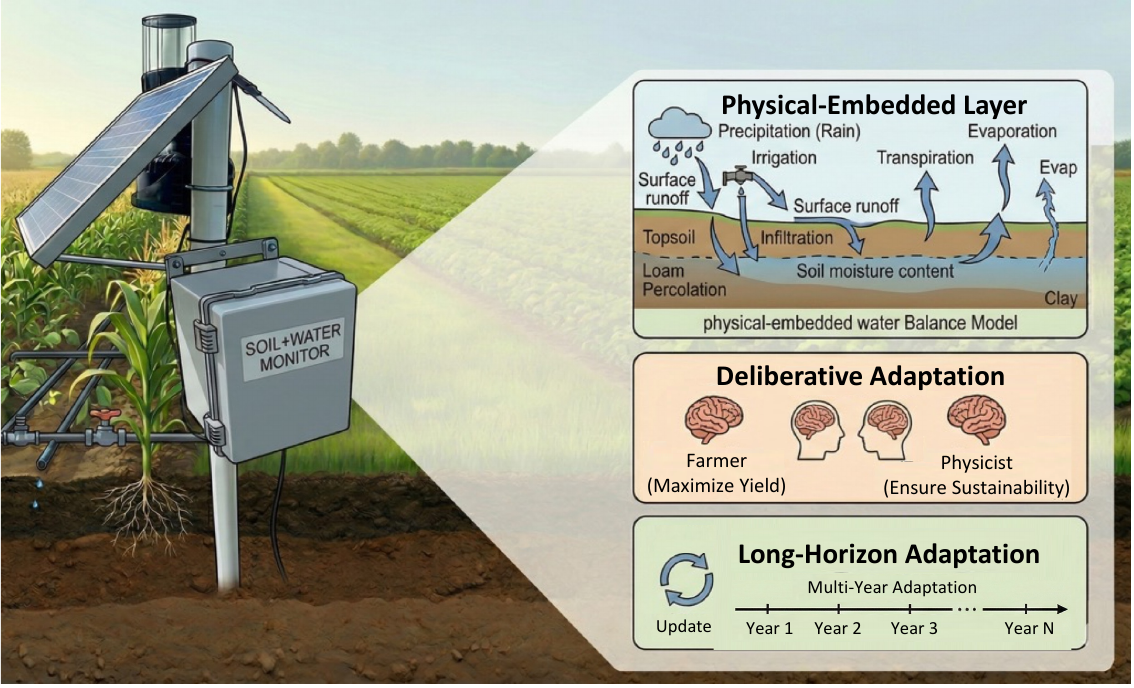}
  \caption{\ours Physical-Embedded Irrigation Control.}
  \label{fig:teaser}
\end{wrapfigure}

This question exposes three coupled challenges. 
\textbf{Q1: physical grounding.} Language models operate naturally over semantic representations, whereas irrigation is governed by quantities with units, conservation relationships, state transitions, and admissible action bounds. 
The agent should not be asked to reconstruct these relationships implicitly from prose. 
\textbf{Q2: decision repair.} 
A physically feasible action is not necessarily a desirable action. 
Multiple feasible actions can induce different delayed stress, drainage, and water-use consequences, so a first language-model proposal should be treated as a hypothesis to verify but not a device command. 
\textbf{Q3: experience-driven adaptation.} Recurrent decision errors may become visible only after repeated operation. 
A persistent agent should learn from those errors, but learning in a physical system raises an additional governance question: what is allowed to change, and what must remain invariant?

We present \ours, a two-timescale, physics-grounded LLM agent that answers these questions by separating \emph{reasoning authority} from \emph{physical authority}. At the fast timescale, a deterministic soil-water model converts observations into a structured physical state and feasible action region. 
The LLM proposes and revises a decision; programmatic simulation exposes the consequences of that proposal; a deterministic selector ranks a bounded candidate set; and runtime assurance retains final execution authority. 
At the slow timescale, recurring corrections and outcome patterns are distilled into persistent contextual principles that can bias future proposals. 
Importantly, this adaptation don't need to modify the transition model, evaluator, action bounds, or execution gate. 
The LLM therefore contributes contextual interpretation, proposal generation, and revision.

Recent continual and self-improving agents increasingly treat memory, prompts, and other framework components as objects that can change from experience without updating model weights~\cite{Kim2024OnlineContinual,Sridhar2026MemER,Ren2026SelfImprovingAgents,Feng2026SelfEvolvingEmbodiedAI}. 
\ours adopts a deliberately bounded version of this idea: experience may repair future reasoning context, but not the trusted physical substrate. 
Fast-loop repair ends when verification no longer changes the proposal or its fixed budget is exhausted; slow-loop memory changes only when recurring evidence crosses a configured trigger. 
We therefore use \emph{bounded adaptation}. 
This boundary distinguishes \ours from language-only refinement~\cite{Madaan2023SelfRefine,du2024debate,Wu2026VIRF} and from broader self-adaptive physical agents~\cite{Kaushik2026PhysicalAI}: language agreement is not physical evidence, memory cannot relax execution constraints, and an LLM proposal never directly becomes an actuator command.

We evaluate \ours retrospectively using field-observed data from Michigan, Texas, and Colorado spanning multiple crops and years. 
In the reported multi-site comparison, \ours attains an aggregate control cost of $100.94\pm1.80$ with $2.58\pm0.12$~mm irrigation under the common evaluator, while the historical schedule replay uses $5.32\pm0.57$~mm. 
Mechanism-specific matched ablations show higher control cost when forward simulation, verified proposal revision, or persistent context is removed. 
A separate nine-season sequence exhibits substantial seasonal variation without sustained divergence.
That shows the ability of long-horizon operational stability. 
Finally, performance across the LLM sizes remains in the same broad range across several model families, indicating that the control architecture surrounding the language model materially shapes behavior.

Our contributions are threefold:
\begin{itemize}[topsep=2pt,itemsep=2pt,parsep=0pt,partopsep=0pt]
    \item \textbf{Physics-grounded agent interface.} We formulate and instantiate an interface in which language reasoning operates over explicit physical state and deterministic transition tools, while state evolution, feasible actions, and execution constraints remain outside the language model.
    \item \textbf{Simulator-grounded decision repair.} We develop a bounded propose-verify-revise mechanism in which an LLM proposal is confronted with numerical rollout evidence before a deterministic selector ranks executable candidates.
    \item \textbf{Bounded experience-driven adaptation.} We integrate evidence-triggered persistent context that can repair future proposal priors from recurring experience while leaving the physical model, evaluator, and runtime constraints immutable.
\end{itemize}

\section{Related Work}
\label{sec:related_work}

\begin{table*}[t]
\centering
\small
\setlength{\tabcolsep}{7pt}
\renewcommand{\arraystretch}{1.12}

\begin{tabularx}{\textwidth}{
    >{\raggedright\arraybackslash}p{4.0cm}
    >{\centering\arraybackslash}X
    >{\centering\arraybackslash}X
    >{\centering\arraybackslash}X
    >{\centering\arraybackslash}X
}
\toprule
\textbf{Representative Paradigm} &
\textbf{Physical Grounding} &
\textbf{Deliberative Reasoning} &
\textbf{Persistent Adaptation} &
\textbf{Independent Runtime Safety} \\
\midrule

Model-based irrigation
\cite{Allen1998FAO56,Delgoda2016IrrigationMPC}
& \checkmark & -- & \(\circ\) & \checkmark \\

Learning-based irrigation
\cite{Li2021RLIrrigation,Ding2024DRLIC}
& \(\circ\) & -- & \(\circ\) & \(\circ\) \\

Embodied LLM agents
\cite{Ahn2022SayCan,Driess2023PalmE,Yao2023ReAct}
& \(\circ\) & \(\circ\) & -- & -- \\

Self-improving agents
\cite{Shinn2023Reflexion,Wang2024Voyager,zhang2025agentic}
& -- & \checkmark & \checkmark & -- \\

Runtime-assured control
\cite{Ames2019CBF,Bak2024BlackBoxSimplex,Wachi2020SafeRL}
& \checkmark & -- & -- & \checkmark \\

\textbf{\ours}
& \checkmark
& \checkmark
& \checkmark
& \checkmark \\

\bottomrule
\end{tabularx}

\caption{
Positioning of \ours relative to representative research paradigms.
\checkmark denotes a primary mechanism, \(\circ\) denotes partial support, and -- denotes that the capability is not a primary focus.
}
\label{tab:related_work}
\end{table*}

\paragraph{Irrigation and long-horizon control.}
Irrigation scheduling is commonly formulated as a physical control problem linking soil-water availability, crop demand, weather forcing, and irrigation actions. FAO-56 provides widely used evapotranspiration and crop-water formulations~\cite{Allen1998FAO56}; MPC optimizes future actions under explicit dynamics and constraints~\cite{Delgoda2016IrrigationMPC,Camacho2013MPC}; and learning-based approaches optimize policies from simulated or observed interaction~\cite{Li2021RLIrrigation,Ding2024DRLIC,Kelly2024AquaCropGym}. 
These approaches are strong control baselines and, when the control problem is fully specified, do not require language reasoning. 
Agricultural LLM systems increasingly incorporate sensor and weather context for irrigation advice or decision support~\cite{Kandamali2025CottonBot,Yu2026IrrigationAgents}. 
Our question is different: how should a language-model agent participate in repeated closed-loop control when numerical prediction and execution authority must remain explicit?

\paragraph{Grounded, embodied, and verifiable LLM agents.}
Grounded agents couple language reasoning to external state, tools, or executable capabilities. SayCan grounds plans in robot affordances~\cite{Ahn2022SayCan}, PaLM-E incorporates embodied observations~\cite{Driess2023PalmE}, and ReAct interleaves reasoning with actions and observations~\cite{Yao2023ReAct}. 
Iterative methods such as Self-Refine and multi-agent debate revisit an initial generation through critique or interaction~\cite{Madaan2023SelfRefine,du2024debate}, while VIRF couples an LLM planner to a deterministic logic tutor for verifiable plan repair~\cite{Wu2026VIRF}. 
\ours shares the principle that generative proposals should be externally checked, but continuous physical control imposes a different contract: numerical simulation determines state consequences, deterministic logic ranks bounded actions, and runtime assurance remains independent of language-model agreement. 

\paragraph{Persistent memory, continual learning, and self-improving agents.}
A growing literature studies agents that accumulate experience. Reflexion stores verbal feedback~\cite{Shinn2023Reflexion}, Voyager accumulates reusable skills~\cite{Wang2024Voyager}, online continual learning has been studied for embodied instruction-following agents~\cite{Kim2024OnlineContinual}, and experience retrieval improves long-horizon robotic control~\cite{Sridhar2026MemER}. 
Agentic Context Engineering (ACE) treats context as an evolving playbook~\cite{zhang2025agentic}. 
More recent work frames changes to prompts, memory, tools, or controller frameworks as forms of agent self-improvement~\cite{Ren2026SelfImprovingAgents}, and proposes broader self-evolving embodied systems that adapt memory, task knowledge, or models over time~\cite{Feng2026SelfEvolvingEmbodiedAI}. 
Concurrent physical-agent work similarly investigates self-adaptive crop management with verification and reflection~\cite{Kaushik2026PhysicalAI}.

\ours occupies a deliberately narrower point in this design space. 
It performs no model-weight update, no online world-model revision, and no autonomous relaxation of safety rules. 
Experience can change only the contextual framework used for future proposals. 
This boundedness is important in physical control because continual adaptation and runtime assurance solve different problems: the former determines what the agent should remember or reconsider, while the latter determines what it is allowed to execute. 
Control barrier functions, Simplex-style runtime assurance, and safe RL provide complementary mechanisms for maintaining explicit safety conditions~\cite{Ames2019CBF,Bak2024BlackBoxSimplex,Wachi2020SafeRL}. 
The contribution of \ours is a two-timescale architecture that permits experience-driven reasoning repair while preserving an immutable numerical control boundary.

\section{Problem Formulation}
\label{sec:problem}

We study persistent physical control in which an agent repeatedly observes an evolving environment, proposes an action, and receives feedback only through subsequent state evolution. 
At decision step $t$, observations $o_t$ are mapped by a fixed physical abstraction to state
\begin{equation}
    s_t=\phi(o_t;\theta_{\mathrm{phys}}),
\end{equation}
and the process evolves as
\begin{equation}
    s_{t+1}=f(s_t,a_t,w_t), \qquad a_t\in\mathcal{A}(s_t),
    \label{eq:problem_dynamics}
\end{equation}
where $w_t$ denotes exogenous forcing and $\mathcal{A}(s_t)$ is the feasible action set. Over horizon $T$, the controller seeks low cumulative physical cost,
\begin{equation}
    J(\pi)=\mathbb{E}\!\left[\sum_{t=1}^{T}c(s_t,a_t)\right],
    \qquad \text{s.t. Eq.~\ref{eq:problem_dynamics}}.
    \label{eq:control_objective}
\end{equation}
Irrigation instantiates this problem with root-zone water state, precipitation, and atmospheric demand as forcing, and irrigation depth as the action.

Our objective is to decompose control into an adaptive semantic layer and a trusted numerical layer. 
Let $x_t$ denote the structured decision context and $m_e$ the persistent context for episode or season $e$. The LLM produces a proposal
\begin{equation}
    z_t=g_{\theta}(x_t,m_e),
    \label{eq:proposal_anchor}
\end{equation}
while the executed action is determined only after physical evaluation and assurance,
\begin{equation}
    a_t=\mathcal{D}(s_t,z_t;f,\mathcal{A}),
    \label{eq:decomposed_policy}
\end{equation}
where $\mathcal{D}$ denotes bounded rollout, candidate selection, runtime assurance, and actuation gating. This decomposition makes the allocation of authority explicit: $g_\theta$ can influence what is proposed or reconsidered, but $f$, $\mathcal{A}$, and $\mathcal{D}$ determine physical consequence and executable action.

Persistent experience acts on a slower timescale. Let $\tau_e$ denote the completed trajectory and outcome summary for episode $e$, and let $q_e$ be the configured evidence trigger extracted from that experience. 
We write
\begin{equation}
    m_{e+1}=
    \begin{cases}
        \mathcal{G}(m_e,\tau_e), & q_e=1,\\
        m_e, & q_e=0,
    \end{cases}
    \label{eq:triggered-memory-update}
\end{equation}
where $\mathcal{G}$ modifies only the contextual framework supplied to future LLM calls. 
In particular, $\theta$, $f$, $\theta_{\mathrm{phys}}$, the evaluator, and runtime constraints are fixed by construction. 
Equation~\ref{eq:triggered-memory-update} formalizes the limited sense in which the agent can improve from experience: the reasoning context can be repaired when recurring evidence warrants an update, but the trusted physical substrate cannot self-modify.

This exposes three questions. \textbf{Q1} How should $x_t$ show physically meaningful state and constraints to language reasoning? \textbf{Q2} how should a proposal $z_t$ be checked and repaired before it can influence $a_t$? \textbf{Q3} how should $m_e$ accumulate experience without changing physical truth or execution authority? 
\ours addresses these questions through a physics-grounded interface, simulator-grounded decision repair, and bounded cross-episode adaptation.

\begin{figure*}[t]
    \centering
    \includegraphics[width=\linewidth]{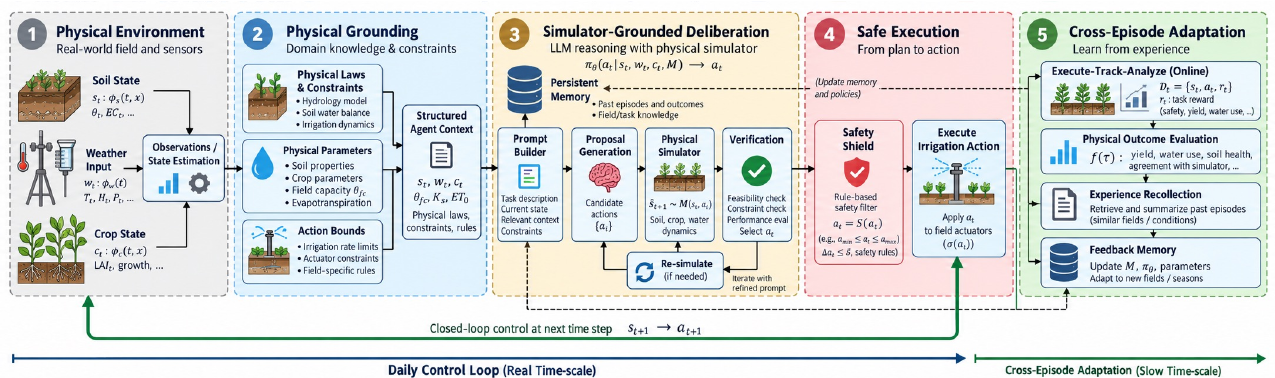}
    \caption{\ours as a two-timescale physical agent. In the daily loop, observations are converted to a structured physical state; the LLM proposes or revises a decision; runtime assurance gates execution. In the slower loop, recurring failures are consolidated into persistent context. Adaptation only changes the reasoning context.}
    \label{fig:system_overview}
\end{figure*}

\section{\ours Design}
\label{sec:mimir-design}

Figure~\ref{fig:system_overview} shows the central design principle: adaptive reasoning and physical authority are intentionally asymmetric. The language model may propose, explain, and revise. 
Numerical tools determine predicted consequences. 
Deterministic logic constrains and ranks actions. 
Runtime assurance retains final actuator authority. 
Persistent memory may influence future reasoning, but it cannot expand what the physical controller permits.
This creates two repair loops: a fast loop corrects individual proposals using physical evidence, while a slow loop consolidates recurring experience into contextual principles. Neither loop can alter physical state-transition rules or execution constraints.

\subsection{Physics-Grounded Agent Interface}
\label{sec:physical_interface}

\ours exposes a structured numerical state. 
Let $S_t$ denote root-zone water storage, $P_t$ precipitation, $ET_t$ evapotranspiration, $I_t$ irrigation, and $D_t$ drainage. 
The controller-side model uses the lightweight balance
\begin{equation}
    S_{t+1}=S_t+P_t+I_t-ET_t-D_t.
    \label{eq:water_balance}
\end{equation}
The reasoning context contains root-zone storage, field capacity ($FC$), wilting point, crop stage, evapotranspiration, and decision-time forecast information. 
This representation is an interface contract in which units, state variables, and admissible quantities are fixed before querying the LLM. 
Appendix~\ref{app:physics} specifies the physical abstraction and its assumptions.

The state also induces an execution-feasible action interval. With anticipated retained precipitation $\widehat P_{\mathrm{eff},t+1}$,
\begin{equation}
    I_t^{\max}=\max\!\left(0,FC-S_t-\widehat P_{\mathrm{eff},t+1}\right),
    \qquad \mathcal{A}_t=[0,I_t^{\max}].
    \label{eq:imax}
\end{equation}
The final command is projected into this interval,
\begin{equation}
    a_t=\Pi_{\mathcal{A}_t}(\widetilde a_t).
    \label{eq:projection}
\end{equation}
This is an upper-bound execution constraint. 
The LLM cannot expand $\mathcal{A}_t$, alter units, or change the transition parameters used by the controller.

\subsection{Simulator-Grounded Decision Repair}
\label{sec:deliberation}

Physical feasibility alone does not determine which feasible action is preferable. 
\ours therefore treats language output as a proposal to be repaired if physical evidence contradicts its rationale. 
Given structured context $x_t$, the proposer generates
\begin{equation}
    \widetilde a_t^{(0)}=\mathrm{Propose}(x_t).
\end{equation}
At repair round $k$, programmatic tools evaluate the proposed consequence,
\begin{equation}
    e_t^{(k)}=\mathrm{Verify}(s_t,\widetilde a_t^{(k)}),
\end{equation}
and the proposal can be revised using the resulting numerical evidence,
\begin{equation}
    \widetilde a_t^{(k+1)}=\mathrm{Revise}(x_t,\widetilde a_t^{(k)},e_t^{(k)}).
    \label{eq:revise}
\end{equation}
The implementation names the proposal and verification roles ``Farmer'' and ``Physicist''.
The distinction is between generative proposal and programmatic verification. 
The fast repair loop terminates when verification no longer changes the proposal or when the fixed round budget is reached, so the agent cannot deliberate indefinitely.

The repaired proposal remains only an \emph{anchor}. 
A deterministic selector constructs a finite set $\mathcal{C}_t\subseteq\mathcal{A}_t$ around the anchor while retaining reference candidates such as zero irrigation and the admissible refill bound. 
Each candidate is rolled out over the control horizon,
\begin{equation}
    \widehat\tau_t(a)=\mathrm{Rollout}(s_t,a,w_{t:t+H}),
\end{equation}
and assigned fixed rollout cost $J_t(a)$. The selector chooses
\begin{equation}
    a_t^\star=\arg\min_{a\in\mathcal{C}_t}
    \left(N_{\mathrm{viol}}(a),J_t(a)\right),
    \label{eq:candidate_selection}
\end{equation}
using lexicographic priority on modeled physical violations and then rollout cost. 
The selected candidate passes runtime assurance and a trusted fallback path before the final projection in Eq.~\ref{eq:projection}.

Language can therefore change the search prior, but not physical evidence or execution authority. The corresponding ablation tests whether numerical feedback improves the LLM-conditioned proposal, not whether language agents replace optimization.

\subsection{Bounded Experience-Driven Adaptation}
\label{sec:adaptation}

The fast loop repairs an individual decision, but persistent agents also encounter repeated failure modes. 
A proposal may repeatedly overreact to forecast uncertainty, require the same verifier correction, or reach the same runtime safeguard across multiple episodes. 
\ours therefore maintains a slower adaptation loop that converts recurring experience into reusable contextual knowledge.

Let $\tau_e$ be the completed trajectory and outcome summary for episode or season $e$. We implement $\mathcal{G}$ in Eq.~\ref{eq:triggered-memory-update} using Agentic Context Engineering~\cite{zhang2025agentic}. 
Deliberation traces, repeated verifier corrections, control-cost patterns, and runtime interventions provide evidence from which we can distill compact principles. 
A principle records the context in which a pattern occurred, the corrective tendency, relevant exceptions, and its supporting evidence. 
When the configured recurrence trigger is not met, the slow loop performs no update; thus adaptation is event-driven rather than an unconditional rewrite after every season.

Memory can condition later proposals or provide a soft anchor, but cannot rank or execute actions. Across updates,
\begin{equation}
    m_e \rightarrow m_{e+1},
\end{equation}
while
\begin{equation}
    g_\theta\text{'s weights},\quad f_{\mathrm{phys}},\quad \theta_{\mathrm{phys}},\quad
    \mathcal{A}_t,\quad \text{Evaluator},\quad \text{Runtime Assurance}
\end{equation}
remain fixed. 
Memory cannot access future held-out outcomes, relax a safety bound, change the scoring rule, or bypass deterministic selection.

We call this \emph{bounded experience-driven adaptation}: framework-level self-improvement in which accumulated experience can change future reasoning context, but not model weights, the physical world model, or runtime constraints. 
The central question is therefore not only whether a physical agent can learn, but \emph{which parts are permitted to learn and under what evidence}. Appendix~\ref{app:ace} gives the full memory contract.

\section{Experimental Setup}
\label{sec:experimental-setup}


\subsection{Datasets and Protocol}
\label{sec:datasets}

The corpus contains approximately 6{,}938 days, or 19 crop-years, spanning four environments: Michigan potato under humid, variable rainfall~\cite{dong2024implementation}; Bushland, Texas maize under high evaporative demand~\cite{bushland_maize}; Colorado LIRF maize under limited-irrigation treatments~\cite{lirf_sunflower}; and a separate Colorado sunflower sequence for the nine-season analysis~\cite{colorado_sunflower}. Available streams include hourly weather forcing, irrigation records, evapotranspiration-related variables, and soil-water observations. Appendix~\ref{app:data} summarizes their role and describes provenance and preprocessing.


For each eligible chronological segment, the first 60\% is reserved for calibration or baseline training and the remaining 40\% for evaluation; samples are never shuffled. Decisions are issued every 24~h using a 168~h history and a 24~h forward horizon, while the physical model integrates hourly. At time $t$, a method can use only information available by that decision time. All methods share the same forecast construction, actuator limits, feasible action bounds, and evaluation window. The protocol forbids future reads, initialization or repair of persistent context from future held-out outcomes, test-based seed selection, and overlap between calibration and evaluation windows. The nine-season sequence preserves chronological order and is analyzed separately from the main aggregate.

\subsection{Reference Controllers and Comparison Contract}
\label{sec:baselines}

We report four reference classes. \textbf{Historical irrigation} replays recorded irrigation actions under the common evaluator and serves as a management reference; it is not a comparison of realized crop yield. \textbf{MAD-50} irrigates after 50\% management-allowable depletion and refills toward field capacity subject to the common action limit~\cite{FAO2017Water}. \textbf{ET-Replace} replaces accumulated evapotranspiration loss using the same causal weather information. \textbf{DRLIC} provides the reinforcement-learning reference~\cite{Ding2024DRLIC}. Appendix~\ref{app:baselines} additionally describes zero irrigation, fixed-threshold control, and a deterministic physics-search/MPC reference used diagnostically to separate language-conditioned anchoring from numerical action search.


\subsection{Metrics}
\label{sec:metrics}

The primary metric is seasonal physical control cost,
\begin{equation}
    C=w_WW+w_SS+w_RR+w_LL,
    \label{eq:eval-cost}
\end{equation}
where $W$ is irrigation applied, $S$ accumulated soil-water stress deficit, $R$ runoff, and $L$ deep percolation/leaching. All terms are computed from the same hourly trajectory and use a single fixed set of weights across sites, controllers, and language-model backbones. Lower $C$ is better. We call this quantity \emph{control cost}.
It is an absolute trajectory objective. 
We separately report irrigation applied in millimeters; lower water use is interpreted only jointly with control cost, because a controller can save water by accepting larger stress. Appendix~\ref{app:metrics} defines the components and aggregation protocol.

\subsection{Implementation Details}
\label{sec:implementation-details}

Unless stated otherwise, \ours uses a DeepSeek-14B backbone served locally through Ollama.
Fast-loop decision repair is capped at three proposal-verification-revision rounds. 
Every proposal is followed by deterministic candidate construction, rollout scoring, runtime assurance, fallback if required, and the final actuation gate; the LLM never writes directly to the actuator.

The slow loop changes only persistent context. 
Recurrence triggers, memory budget, activation/reset rules, and the representation of stored principles are fixed by configuration within each matched experiment. 
A season with no qualifying recurring pattern produces no contextual update. 
For model-scale and model-family studies, only the language-model checkpoint changes; prompts, physical state, simulator, action bounds, repair budget, candidate selection, and evaluation protocol remain fixed. 
Appendix~\ref{app:llm} reports prompting and parsing details, and Appendix~\ref{app:ace} details the persistent-context update contract.

\section{Evaluation}
\label{sec:evaluation}

In the evaluation We first ask whether the complete controller attains useful long-horizon performance under a common protocol. 
We then test the three proposed capabilities in order: physical grounding (Q1), verified decision repair (Q2), and experience-driven adaptation (Q3). 
Because several component ablations were produced on different matched evaluation slices, each effect is interpreted only against the corresponding full configuration.

\subsection{Multi-Site and Multi-Year Control Performance}
\label{sec:rq1}

Table~\ref{tab:overall-results} reports the multi-site aggregate. \ours obtains control cost $100.94\pm1.80$ with $2.58\pm0.12$~mm irrigation. MAD-50 records $134.33\pm3.10$ at $10.58\pm0.44$~mm, and ET-Replace records $440.74\pm12.48$ at $12.70\pm0.31$~mm. DRLIC applies the least water ($0.35\pm0.00$~mm) but incurs the largest reported cost ($924.47\pm27.50$), illustrating why water use cannot be interpreted independently of stress and loss terms. Historical irrigation replay uses $5.32\pm0.57$~mm and has cost $790.51\pm17.76$ under the same evaluator. Within this reported set, \ours therefore occupies the lowest-cost operating point; this comparison is not a claim of dominance over all model-based irrigation controllers.

\begin{figure}[!t]
    \centering
    \begin{subfigure}[b]{0.26\linewidth}
        \centering
        \includegraphics[width=\linewidth]{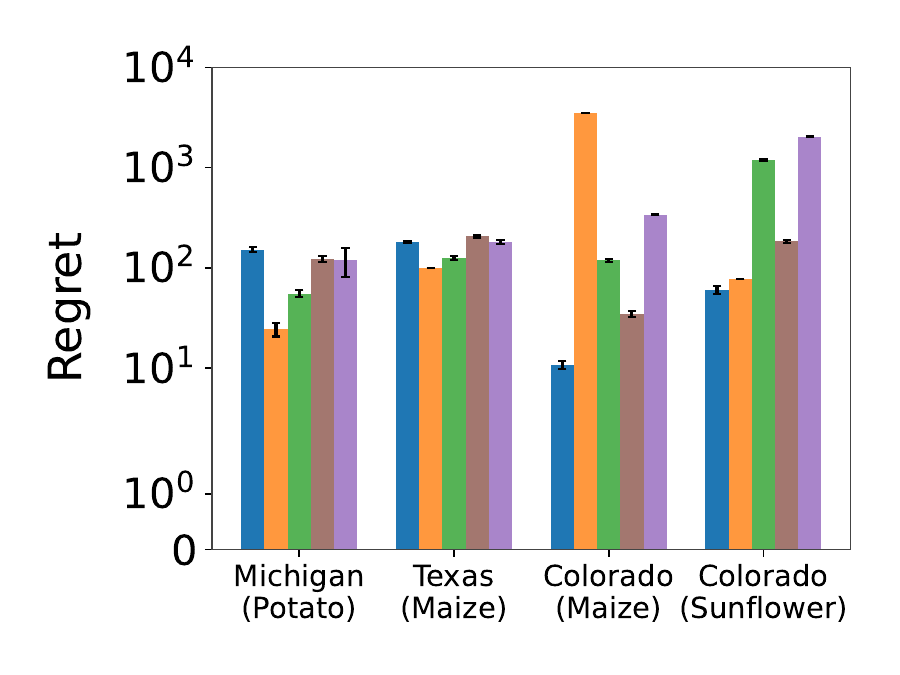}
        \caption{Regional control cost.}
        \label{fig:regional-regret}
    \end{subfigure}
    \hfill
    \begin{subfigure}[b]{0.26\linewidth}
        \centering
        \includegraphics[width=\linewidth]{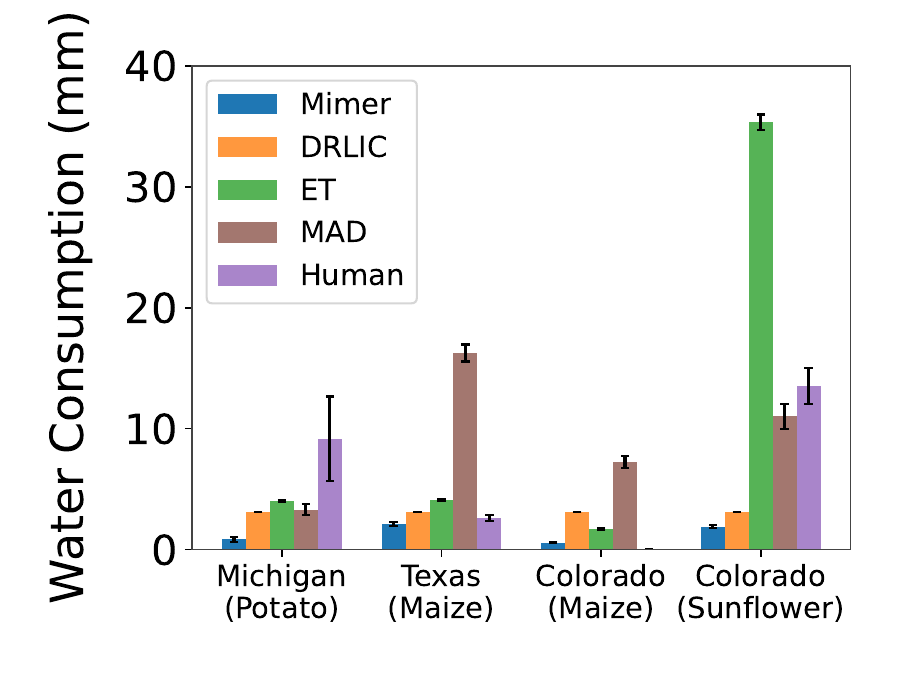}
        \caption{Regional irrigation.}
        \label{fig:regional-irrigation}
    \end{subfigure}
        \centering
    \begin{subfigure}[b]{0.23\linewidth}
        \centering
        \includegraphics[width=\linewidth]{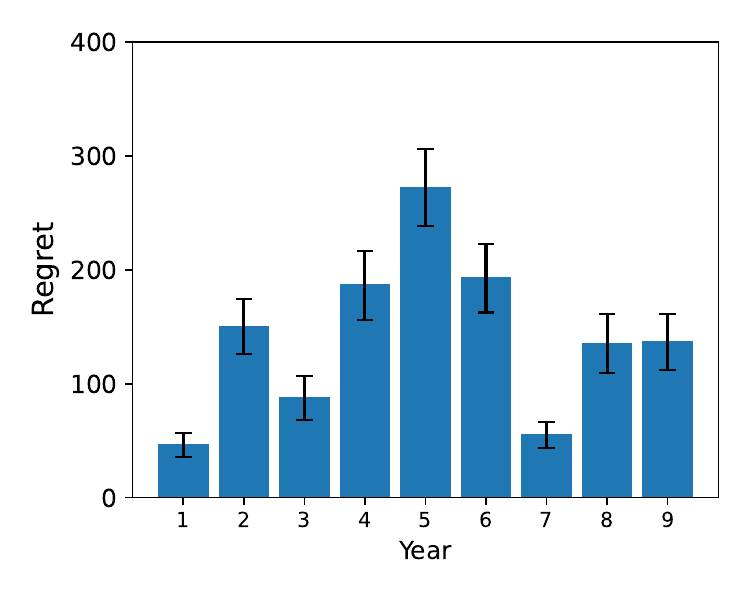}
        \caption{Annual control cost.}
        \label{fig:long-horizon-regret}
    \end{subfigure}
    \hfill
    \begin{subfigure}[b]{0.23\linewidth}
        \centering
        \includegraphics[width=\linewidth]{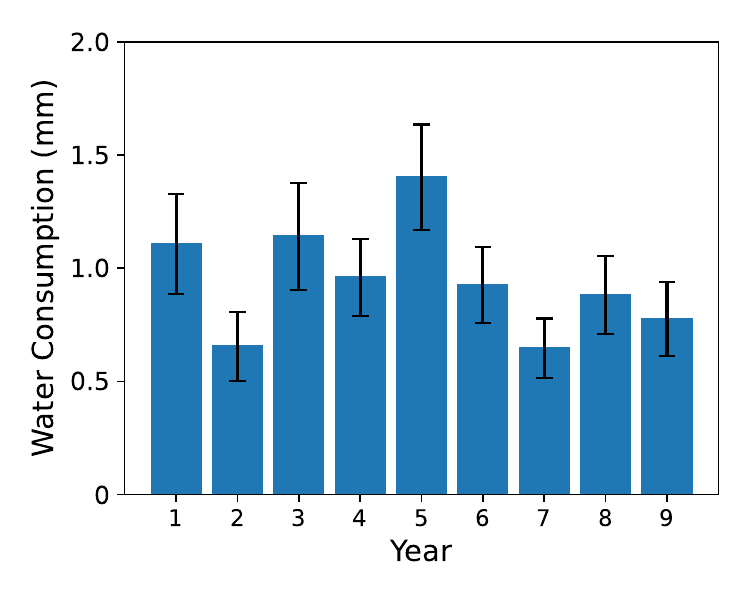}
        \caption{Annual irrigation.}
        \label{fig:long-horizon-irrigation}
    \end{subfigure}
    \caption{Regional results for Michigan potato, Texas maize, Colorado maize, and Colorado sunflower environments. Nine-season Colorado sequence. This experiment evaluates operational stability, not monotonic self-improvement.}
\end{figure}

\begin{table}[t]
    \centering
    \small
    \setlength{\tabcolsep}{6pt}
    \caption{Aggregate performance under the common multi-site evaluation protocol. Control cost is the primary objective; irrigation is reported separately because lower water alone can correspond to higher stress.}
    \label{tab:overall-results}
    \begin{tabular}{lcc}
        \toprule
        \textbf{Controller} & \textbf{Control cost $\downarrow$} & \textbf{Irrigation (mm)} \\
        \midrule
        \ours & $\mathbf{100.94 \pm 1.80}$ & $2.58 \pm 0.12$ \\
        MAD-50 & $134.33 \pm 3.10$ & $10.58 \pm 0.44$ \\
        ET-Replace & $440.74 \pm 12.48$ & $12.70 \pm 0.31$ \\
        DRLIC & $924.47 \pm 27.50$ & $0.35 \pm 0.00$ \\
        Historical irrigation & $790.51 \pm 17.76$ & $5.32 \pm 0.57$ \\
        \bottomrule
    \end{tabular}
\end{table}

The regional view in Fig.~\ref{fig:regional-regret} and Fig.~\ref{fig:regional-irrigation} checks whether the aggregate behavior is concentrated in one environment. \ours records $152.25\pm5.91$ cost and $1.54\pm0.38$~mm irrigation in Michigan, $129.16\pm2.71$ and $3.80\pm0.31$~mm in Texas, and $59.82\pm4.07$ and $3.43\pm0.19$~mm in the reported Colorado evaluation. Because horizon length and forcing differ across crops and sites, we do not use raw regional cost to rank environments; the figure is used only to assess whether the same controller remains viable under distinct environmental conditions.

\subsection{Performance of Explicit Physical Grounding}
\label{sec:rq2}

We ablate two distinct responsibilities of the physical substrate: predicting delayed consequences and enforcing executable bounds. Removing forward simulation increases control cost from $149.5$ to $730.5$ on its matched slice, while irrigation falls from $3.29$ to $0.45$~mm. The joint decrease in water and increase in cost is consistent with under-irrigation whose delayed deficit is not represented during proposal evaluation. Removing the execution shield increases cost from $147.2$ to $259.9$ while mean irrigation changes comparatively little. The second result indicates a different role: execution bounding can alter costly tail behavior without determining the controller's mean irrigation level. Together, the two comparisons support the bounded claim that explicit physical prediction and execution constraints materially affect behavior under the tested conditions. Neither comparison guarantees future agronomic safety. The direct/zero-shot LLM diagnostic used a different metric configuration and is therefore reported only in Appendix~\ref{app:ablations}; its scalar value is not combined with the matched ablations.

\subsection{Performance of Simulator-Grounded Decision Repair}
\label{sec:rq3}

We remove the proposal--verification--revision stage while retaining the downstream simulator, candidate selector, and execution constraints. On the matched slice, control cost increases from $109.1$ to $180.5$ and irrigation from $2.47$ to $3.19$~mm. Because the two variants share the same downstream physical substrate, the difference isolates the value of revising the language-conditioned anchor using explicit numerical feedback. The result supports a narrow conclusion: verified repair improves the evaluated objective on this slice relative to the corresponding single-pass proposal. It does not establish that free-form multi-agent debate is superior to every alternative inference-time strategy, nor that the language model replaces deterministic search.

\begin{table}[t]
    \centering
    \small
    \setlength{\tabcolsep}{4.2pt}
    \caption{Paired component ablations. Each row has its own matched full configuration; absolute full values differ across slices and must not be compared between rows. Water is in mm.}
    \label{tab:paired-ablations}
    \begin{tabular}{lcccc}
        \toprule
        & \multicolumn{2}{c}{\textbf{Full}} & \multicolumn{2}{c}{\textbf{Ablated}} \\
        \cmidrule(lr){2-3}\cmidrule(lr){4-5}
        \textbf{Removed component} & \textbf{Cost} & \textbf{Water} & \textbf{Cost} & \textbf{Water} \\
        \midrule
        Persistent context (ACE) & 103.0 & 3.40 & 157.5 & 4.41 \\
        Decision repair & 109.1 & 2.47 & 180.5 & 3.19 \\
        Execution shield & 147.2 & -- & 259.9 & $\approx$ unchanged \\
        Forward simulator & 149.5 & 3.29 & 730.5 & 0.45 \\
        \bottomrule
    \end{tabular}
\end{table}

\subsection{Performance of Cross-Year Experience}
\label{sec:rq4}

Removing persistent ACE context increases control cost from $103.0$ to $157.5$ and irrigation from $3.40$ to $4.41$~mm on its matched slice (Table~\ref{tab:paired-ablations}). The physical simulator, candidate selector, and execution shield are unchanged. The comparison therefore supports the statement that information carried across episodes improves later proposal quality beyond the instantaneous physical state under this protocol. It is evidence for useful experience accumulation, not for monotonic continual learning.

We separately examine nine consecutive Colorado seasons. Figure~\ref{fig:long-horizon-regret} and Figure~\ref{fig:long-horizon-irrigation} shows substantial year-to-year variation, including a maximum reported annual cost of $272.15$ in Year~5, followed by lower values rather than sustained divergence. We interpret this experiment as evidence of operational stability across changing seasonal forcing. It does not establish that the agent improves every year, that every memory update is beneficial, or that the memory mechanism is immune to forgetting. Those stronger properties would require dedicated continual-learning evaluation beyond the current experiments.

\section{Conclusion}
\label{sec:conclusion}

We presented \ours, a physics-grounded LLM agent for long-horizon irrigation control. 
The design begins from a simple premise: persistent physical agents need the ability to adapt, but not every part of a physical controller should be equally free to change. 
\ours therefore separates three responsibilities. 
A structured physical interface defines state and feasible actions; 
simulator-grounded decision repair lets language-model proposals be challenged with numerical evidence before deterministic selection; 
and bounded cross-episode context carries recurring experience forward without changing the physical model or execution constraints. 
Across the reported multi-site and multi-year evaluation, matched ablations show that physical simulation, verified repair, and persistent context each affect control behavior. 
The resulting contribution shows one way to build a persistent physical agent in which reasoning can evolve from experience while physical truth and actuator authority remain explicitly governed.


\clearpage
\appendix


\section*{Appendix Overview}
\phantomsection
\label{sec:appendix_overview}

This appendix provides additional methodological, implementation,
evaluation, and reproducibility details supporting the main paper.
Each entry below is hyperlinked to the corresponding section.

\begin{center}
\footnotesize
\setlength{\tabcolsep}{4.5pt}
\renewcommand{\arraystretch}{1.08}
\begin{tabularx}{\textwidth}{
    >{\centering\arraybackslash}p{0.65cm}
    >{\raggedright\arraybackslash}p{4.75cm}
    >{\raggedright\arraybackslash}X
}
\toprule
\textbf{Sec.} & \textbf{Content} & \textbf{Purpose} \\
\midrule
\hyperref[app:physical_model]{A} &
\hyperref[app:physical_model]{\textbf{Physical Model and Execution Constraints}} &
Defines the physical state, root-zone water balance, feasible actions,
and execution-time constraints. \\

\hyperref[app:datasets]{B} &
\hyperref[app:datasets]{\textbf{Dataset Provenance and Preprocessing}} &
Describes the multi-site datasets, preprocessing, temporal resolution,
unit normalization, and causal data handling. \\

\hyperref[app:evaluation_protocol]{C} &
\hyperref[app:evaluation_protocol]{\textbf{Evaluation Protocol and Leakage Controls}} &
Details chronological splitting, causal-information constraints,
matched comparisons, and safeguards against evaluation leakage. \\

\hyperref[app:metrics]{D} &
\hyperref[app:metrics]{\textbf{Metric Definition}} &
Specifies the seasonal physical control cost, component metrics,
aggregation, and uncertainty reporting. \\

\hyperref[app:baselines]{E} &
\hyperref[app:baselines]{\textbf{Reference Controller Implementations}} &
Documents historical, heuristic, model-based, and learning-based
irrigation references under the common evaluation interface. \\

\hyperref[app:llm]{F} &
\hyperref[app:llm]{\textbf{Language-Model Interface}} &
Reports the LLM configuration, structured prompting interface,
decoding settings, parsing, fallback behavior, and telemetry. \\

\hyperref[app:decision_repair]{G} &
\hyperref[app:decision_repair]{\textbf{Simulator-Grounded Decision Repair}} &
Provides the proposal--verification--revision loop, bounded candidate
search, deterministic selection, and a representative trace. \\

\hyperref[app:adaptation]{H} &
\hyperref[app:adaptation]{\textbf{Bounded Experience-Driven Adaptation}} &
Describes persistent contextual memory, evidence-triggered updates,
stopping conditions, and immutable adaptation boundaries. \\

\hyperref[app:full_results]{I} &
\hyperref[app:full_results]{\textbf{Full Quantitative Results}} &
Reports aggregate and regional results underlying the main-paper comparisons. \\

\hyperref[app:ablations]{J} &
\hyperref[app:ablations]{\textbf{Component Ablations}} &
Provides matched component ablations and additional diagnostics. \\

\hyperref[app:model_robustness]{K} &
\hyperref[app:model_robustness]{\textbf{Foundation-Model Robustness}} &
Reports sensitivity to language-model scale and model family under
an otherwise fixed physical-control architecture. \\

\hyperref[app:long_horizon]{L} &
\hyperref[app:long_horizon]{\textbf{Nine-Season Long-Horizon Analysis}} &
Provides year-by-year results used to assess long-horizon operational stability. \\

\hyperref[app:reproducibility]{M} &
\hyperref[app:reproducibility]{\textbf{Reproducibility and Provenance}} &
Summarizes configuration freezing, experiment provenance, artifact logging,
and reproduction requirements. \\

\hyperref[app:discussion]{N} &
\hyperref[app:discussion]{\textbf{Discussion and Limitations}} &
Clarifies claim scope, adaptation boundaries, evidence limitations,
and implications for persistent physical agents. \\

\hyperref[app:ai_use]{O} &
\hyperref[app:ai_use]{\textbf{AI Use Disclosure}} &
Discloses the use and verification of generative AI tools in the research workflow. \\
\bottomrule
\end{tabularx}
\end{center}

\section{Physical Model and Execution Constraints}
\label{app:physical_model}
\label{app:physics}

\subsection{Root-Zone State and Units}
All water quantities are represented as equivalent water depth in millimeters.
When volumetric soil moisture $\theta_t$ (m$^3$/m$^3$) is available, the
controller maps it to root-zone storage as
\begin{equation}
    S_t = \theta_t Z_{\mathrm{root}},
    \label{eq:app-root-zone}
\end{equation}
with $Z_{\mathrm{root}}=600\,\mathrm{mm}$ in the reported configuration.
This fixed depth is part of the controller abstraction.

The structured controller state contains root-zone storage $S_t$, field
capacity $FC$, wilting point $WP$, crop-stage information, evapotranspiration,
observed precipitation, and the decision-time precipitation input. We keep
three quantities conceptually separate throughout the system:
(i) physical storage limits, (ii) the agronomic stress threshold used by the
evaluator, and (iii) the feasible action bound used by the controller.

\subsection{Water-Balance Transition}
The controller-side model evolves root-zone storage according to
\begin{equation}
    S_{t+1} = S_t + P_t + I_t - ET_t - D_t,
    \label{eq:app-water-balance}
\end{equation}
where $P_t$, $I_t$, $ET_t$, and $D_t$ denote precipitation, irrigation,
evapotranspiration, and drainage/deep percolation, respectively. The bucket
approximation uses
\begin{equation}
    D_t =
    \max\!\left(
        0,\,
        S_t + P_t + I_t - ET_t - FC
    \right).
    \label{eq:app-drainage}
\end{equation}

The model is intentionally lightweight. It is used to construct and verify bounded irrigation decisions.
Reference evapotranspiration is computed from the available meteorological
forcing using FAO-56 Penman--Monteith where the required variables are
available~\cite{Allen1998FAO56}; otherwise, the implementation follows a
deterministic fallback path fixed before evaluation and shared by all
controllers.

\subsection{Effective Precipitation and Causal Forecast Input}
The controller can map forecast precipitation to retained root-zone input using
the fixed transform
\begin{equation}
P_{\mathrm{eff}} =
\begin{cases}
P\frac{125-0.2P}{125}, & P \leq 250,\\
125+0.1P, & P > 250.
\end{cases}
\label{eq:app-effective-precip}
\end{equation}
Here $P$ denotes the precipitation quantity supplied through the controller's
forecast interface, expressed in millimeters for the corresponding decision
interval. We treat Eq.~\ref{eq:app-effective-precip} strictly as a fixed
implementation heuristic in the controller interface. Every controller receives the same
transformed causal input, and no controller may access realized future rainfall
beyond the frozen forecast interface.

\subsection{Feasible Action and Execution Gate}
Before actuation, the controller computes the refill upper bound
\begin{equation}
    I_t^{\max}
    =
    \max\!\left(
        0,\,
        FC-S_t-\widehat{P}_{\mathrm{eff},t+1}
    \right),
    \label{eq:app-irrigation-cap}
\end{equation}
which induces the legal action interval
\begin{equation}
    \mathcal{A}(S_t) = [0,I_t^{\max}].
\end{equation}
Given a proposal $\widetilde I_t$, the final actuation gate applies
\begin{equation}
    I_t
    =
    \Pi_{\mathcal{A}(S_t)}(\widetilde I_t)
    =
    \min\!\left(
        \max(0,\widetilde I_t),
        I_t^{\max}
    \right).
    \label{eq:app-actuation-gate}
\end{equation}

This mechanism provides execution-feasibility and upper-bound protection.
It does not guarantee avoidance of future crop-water deficit under arbitrary
forecast error. Mass-balance checks and the device-level fallback/watchdog path
remain external to the LLM and cannot be modified by persistent context.


\section{Dataset Provenance and Preprocessing}
\label{app:datasets}
\label{app:data}

\begin{table}[t]
    \centering
    \small
    \setlength{\tabcolsep}{4pt}
    \caption{Dataset inventory used by the evaluation protocol.}
    \label{tab:app-datasets}
    \begin{tabular}{lllll}
        \toprule
        \textbf{Environment} & \textbf{Crop} & \textbf{Resolution} &
        \textbf{Sequence} & \textbf{Use} \\
        \midrule
        Michigan/LOCOSMOS & Potato & Hourly & 2021--2023 entries & Main evaluation \\
        Bushland, Texas & Maize & Hourly & Multi-year & Main evaluation \\
        Colorado LIRF & Maize & Hourly & 5-year sequence & Main evaluation \\
        Colorado sunflower & Sunflower & Hourly & 9-season sequence & Long-horizon analysis \\
        \bottomrule
    \end{tabular}
\end{table}

The combined corpus contains approximately 6{,}938 days, corresponding to
roughly 19 crop-years. The environments span humid, semi-arid, and continental
agricultural conditions and include multiple crop systems. All input streams are
converted to canonical units before entering the controller. Hourly weather and
soil trajectories are retained for physical integration, whereas control
decisions are issued at daily boundaries.

Missing values are handled before window construction. No imputation step may
use observations that occur after the current decision time. Soil parameters are
calibrated only on the calibration portion of each chronological sequence and
are then frozen for held-out evaluation. Crop-stage and weather variables exposed
to a controller are restricted to information available at the corresponding
decision time.

The nine-season Colorado sunflower sequence is retained in chronological order
for long-horizon analysis and is not pooled into the primary multi-site held-out
aggregate.


\section{Evaluation Protocol and Leakage Controls}
\label{app:evaluation_protocol}
\label{app:protocol}

\subsection{Chronological Split}
For the main comparison, the first 60\% of each eligible chronological segment
is reserved for calibration or baseline training and the remaining 40\% for
evaluation. Time series are never shuffled. Each control decision uses a
168-hour history and a 24-hour control horizon. The physical process is
integrated at hourly resolution, and the next controller invocation occurs
after 24 hours.

The nine-season sequence is analyzed as a preserved chronological sequence.

\subsection{Causal Information Policy}
The evaluation protocol enforces the following leakage guards:
\begin{itemize}[topsep=2pt,itemsep=1pt]
    \item no controller may access observations after the current decision time;
    \item persistent context may not be initialized or updated from future held-out outcomes;
    \item held-out results may not be used to select a favorable random seed;
    \item calibration/training and evaluation windows do not overlap.
\end{itemize}

The forecast interface is identical for all methods. In retrospective replay,
all controllers receive the same causal precipitation input and the same
precipitation-persistence construction where that forecast rule is used.

\subsection{Matched Controller and Component Comparisons}
For every controlled comparison, the initial state, weather sequence, observation
history, physical action range, metric definition, and random seed are held
fixed. Changing the controller does not change the environmental forcing or the
evaluation rule.

Each ablation is interpreted only against the full configuration evaluated on
the same matched slice. Several ablations were produced on different matched
slices, so their absolute full-system values differ. Consequently, ablation
effect sizes are interpreted within row and are not ranked across rows.


\section{Metric Definition}
\label{app:metrics}

\subsection{Seasonal Physical Control Cost}
The frozen evaluator computes
\begin{equation}
    C = w_W W + w_S S + w_R R + w_L L,
    \label{eq:app-control-cost}
\end{equation}
where
\begin{align}
    W &= \sum_t I_t,\\
    S &= \sum_t \max(0,S_{\mathrm{stress}}-S_t),\\
    R &= \sum_t R_t,\\
    L &= \sum_t L_t.
\end{align}

Here $W$ is irrigation applied, $S$ is accumulated storage deficit below the
configured stress threshold, $R$ is runoff, and $L$ is deep
percolation/leaching. All components are computed from the same hourly
trajectory. The weights $(w_W,w_S,w_R,w_L)$ are fixed globally across sites,
controllers, and language-model backbones.

Because Eq.~\ref{eq:app-control-cost} is an absolute trajectory objective.

\paragraph{Frozen numerical weights.}
The exact numerical values of $(w_W,w_S,w_R,w_L)$ are taken from the frozen
experiment configuration associated with the reported results. They should be
inserted here verbatim from that configuration in the final archival version.

\subsection{Irrigation Applied}
We separately report
\begin{equation}
    W = \sum_t I_t
\end{equation}
in millimeters. Lower irrigation is not interpreted as better control in
isolation because a zero-action controller can minimize water use while
accumulating substantial stress-related cost.

\subsection{Aggregation and Uncertainty}
Tables report mean and standard deviation over the repeated evaluation units
defined by the corresponding frozen experiment configuration. The statistical
unit and sample count $n$ are retained with the result artifact for each
table/figure and should be reported directly from that artifact in the final
archival version. No claim of statistical significance is made without an
explicit hypothesis test or confidence interval.


\section{Reference Controller Implementations}
\label{app:baselines}

\subsection{Historical Irrigation Replay}
The historical reference replays recorded irrigation actions under the same
evaluator used for algorithmic controllers. It is included as a practical
management reference and is not interpreted as a comparison to realized grower
yield.

\subsection{Zero Irrigation and Fixed Threshold}
A zero-action controller provides a lower-water diagnostic for identifying
degenerate water-saving behavior. A fixed-threshold controller irrigates from
the current soil-water state using a pre-specified threshold and has no learned
or language-model state.

\subsection{MAD-50}
MAD-50 computes available water from field capacity and the configured lower
agronomic threshold and triggers irrigation after 50\% allowable depletion.
When triggered, it refills toward field capacity subject to the shared actuator
limit. It receives no privileged future information.

\subsection{ET Replacement}
ET-Replace accumulates evapotranspiration loss since the previous irrigation
event and requests replacement of that loss, clipped to the same feasible
action interval as the other controllers. Weather input and ET computation
follow the same causal interface used by \ours.

\subsection{Deterministic Physics-Search/MPC Diagnostic}
A deterministic physics-search controller enumerates a finite candidate action
set under the same controller-side physical model and selects the candidate
with the minimum rollout objective. We use this controller diagnostically to
distinguish language-conditioned anchoring from numerical action search. 

\subsection{DRLIC and Learning References}
The reproduced DRLIC reference uses the common state/action interface and the
same physical action limits as \ours in the reported comparison. Learning-based
values are treated as comparable only when the trained policy is replayed
through the common evaluator; externally reported rewards are not mixed with
the manuscript's control-cost metric.


\section{Language-Model Interface}
\label{app:llm}

The main configuration uses a frozen DeepSeek-14B model through a local Ollama
client. The prompt contains the structured physical state, actuator limits,
forecast information, and active persistent principles and requests structured
JSON output.

The primary decoding configuration uses greedy decoding with temperature zero
where applicable, a fixed software seed, \texttt{think=False},
\texttt{format=json}, and \texttt{num\_predict=2048}. No model-weight update,
fine-tuning, or gradient step occurs during control or across seasons.

Only designated action fields are parsed from the model response. Malformed
responses enter a deterministic fallback path and cannot bypass candidate
selection, runtime assurance, or the final actuation gate. Per-call telemetry
records success/error state, retry count, latency, prompt length, response
length, and model identifier. Experiment-level metadata additionally record the
software revision and data hash.


\section{Simulator-Grounded Decision Repair}
\label{app:decision_repair}
\label{app:deliberation}
\label{app:algorithm}

\subsection{Proposal--Verification--Revision Loop}
At decision time, \ours executes the bounded repair process
\begin{align}
    a_t^{(0)} &= \mathrm{Propose}(x_t),\\
    e_t^{(k)} &= \mathrm{Verify}(s_t,a_t^{(k)}),\\
    a_t^{(k+1)} &= \mathrm{Revise}(x_t,a_t^{(k)},e_t^{(k)}).
\end{align}
The loop terminates when physical verification no longer changes the proposed
action or after the fixed three-round budget.

``Farmer'' and ``Physicist'' are implementation role names. The former proposes
an irrigation action and rationale; the latter invokes numerical tools and
returns structured physical feedback. The methodological contribution is the
separation of proposal generation from programmatic physical verification, not
the role names themselves or free-form debate in isolation.

\subsection{Bounded Candidate Selection}
The deliberation output does not directly determine the executed action. Instead,
it provides an anchor for a deterministic candidate selector.

Without a language-conditioned anchor, the selector constructs candidates over
the legal action interval and retains physically meaningful references,
including zero irrigation and refill toward field capacity. With an LLM/ACE
anchor, the selector searches a bounded neighborhood around that anchor while
retaining the same reference actions. A large raw LLM output cannot enlarge the
legal action interval.

Each daily candidate is expanded over the 24-hour horizon, rolled out by the
fixed controller-side model, and evaluated by the frozen rollout objective.
Selection minimizes
\begin{equation}
    \left(
        N_{\mathrm{viol}}(a),
        J(a)
    \right),
    \label{eq:app-lexicographic}
\end{equation}
lexicographically, where $N_{\mathrm{viol}}$ is the modeled
physical-violation count and $J$ is the rollout cost. The selected action is
subsequently passed to runtime assurance, fallback logic if required, and the
final actuation gate.

The fast-loop execution sequence is therefore:
\[
\text{proposal}
\rightarrow
\text{verification/revision}
\rightarrow
\text{anchor}
\rightarrow
\text{bounded search}
\rightarrow
\text{physical rollout}
\rightarrow
\text{selection}
\rightarrow
\text{runtime assurance}
\rightarrow
\text{actuation gate}.
\]

\subsection{Representative Trace}
A representative trace illustrates the interaction:
\begin{quote}
\small
State: $S_t=19.7\,\mathrm{mm}$, $WP=10.5\,\mathrm{mm}$,
$FC=71.8\,\mathrm{mm}$, $ET_t=3.8\,\mathrm{mm}$,
$\widehat P_{\mathrm{eff},t+1}=0\,\mathrm{mm}$.\\
Proposal: $8.0\,\mathrm{mm}$.\\
Verification: the simulated next state remains above the configured lower
threshold; additional irrigation raises modeled drainage risk.\\
Revision: $4.0\,\mathrm{mm}$.\\
Verification: the zero-irrigation rollout remains feasible under the same
short-horizon model.\\
Final revision: $0.0\,\mathrm{mm}$.
\end{quote}
This trace is illustrative only. Quantitative conclusions rely on matched
component comparisons.


\section{Bounded Experience-Driven Adaptation}
\label{app:adaptation}
\label{app:ace}

The slow loop modifies the contextual framework supplied to future language-model
calls without changing model weights or the physical controller. Let $\tau_e$
denote the completed trajectory and outcome summary for episode/season $e$,
$m_e$ the persistent context, and $q_e$ a configuration-controlled
recurrence/evidence trigger. The update takes the form
\begin{equation}
    m_{e+1}
    =
    \begin{cases}
        \mathcal{G}(m_e,\tau_e), & q_e=1,\\
        m_e, & q_e=0.
    \end{cases}
    \label{eq:app-memory-update}
\end{equation}

The system is not forced to rewrite its context after every episode: if no
recurring pattern satisfies the configured evidence condition, no contextual
update occurs.

ACE implements $\mathcal{G}$ as a soft principle/anchor memory used by the
reasoning layer. Deliberation traces, repeated verifier corrections,
control-cost patterns, and runtime-assurance interventions can provide evidence
for a candidate principle. A stored principle contains an identifier, source
context, trigger, suggested action tendency, exception, supporting evidence,
version, and activation state.

The adaptation pathway preserves the following invariants:
\begin{itemize}[topsep=2pt,itemsep=1pt]
    \item no language-model weight update;
    \item no modification of physical parameters, transition equations, or evaluation rules;
    \item no relaxation of action or runtime-assurance bounds;
    \item no future held-out observations in persistent context;
    \item no bypass of candidate selection, runtime assurance, fallback, or the actuation gate.
\end{itemize}

Accordingly, ``adaptation'' denotes a change in future reasoning context. This is
scaffold/framework-level.

\subsection{Stopping Conditions}
At the fast timescale, proposal repair stops when physical verification no
longer changes the proposal or the fixed three-round budget is reached. At the
slow timescale, no persistent memory update occurs unless the
recurrence/evidence trigger is satisfied.

\paragraph{Frozen adaptation configuration.}
Memory-update thresholds, recurrence rules, memory budget, and reset policy are
fixed within each matched comparison and recorded in the corresponding frozen
experiment configuration. Their exact values should be copied verbatim from the
frozen configuration into the final archival supplement.


\section{Full Quantitative Results}
\label{app:full_results}
\label{app:full-results}

\begin{table}[t]
    \centering
    \small
    \caption{
        Aggregate results under the common multi-site evaluator.
        Irrigation is reported separately from control cost.
    }
    \label{tab:app-overall}
    \begin{tabular}{lcc}
        \toprule
        \textbf{Controller} & \textbf{Control cost} & \textbf{Irrigation (mm)} \\
        \midrule
        \ours & $100.94\pm1.80$ & $2.58\pm0.12$ \\
        MAD-50 & $134.33\pm3.10$ & $10.58\pm0.44$ \\
        ET-Replace & $440.74\pm12.48$ & $12.70\pm0.31$ \\
        DRLIC & $924.47\pm27.50$ & $0.35\pm0.00$ \\
        Historical irrigation & $790.51\pm17.76$ & $5.32\pm0.57$ \\
        \bottomrule
    \end{tabular}
\end{table}

The explicitly reported regional \ours values are:
Michigan, $152.25\pm5.91$ control cost and
$1.54\pm0.38\,\mathrm{mm}$ irrigation;
Texas, $129.16\pm2.71$ and $3.80\pm0.31\,\mathrm{mm}$; and
Colorado, $59.82\pm4.07$ and $3.43\pm0.19\,\mathrm{mm}$.

Absolute control-cost magnitudes are not interpreted as directly comparable
across crops or sites when the evaluated horizons and environmental forcings
differ. The regional results are used to assess consistency across heterogeneous
environments.


\section{Component Ablations}
\label{app:ablations}

\begin{table}[t]
    \centering
    \small
    \caption{
        Paired component ablations. Each row uses its own matched evaluation
        slice; effect sizes must not be compared across rows. Water is in mm.
    }
    \label{tab:app-ablations}
    \begin{tabular}{lcccc}
        \toprule
        & \multicolumn{2}{c}{\textbf{Full}} &
        \multicolumn{2}{c}{\textbf{Ablated}} \\
        \cmidrule(lr){2-3}\cmidrule(lr){4-5}
        \textbf{Removed component} &
        \textbf{Cost} & \textbf{Water} &
        \textbf{Cost} & \textbf{Water} \\
        \midrule
        Persistent context (ACE) & 103.0 & 3.40 & 157.5 & 4.41 \\
        Decision repair & 109.1 & 2.47 & 180.5 & 3.19 \\
        Execution shield & 147.2 & -- & 259.9 & $\approx$ unchanged \\
        Forward simulator & 149.5 & 3.29 & 730.5 & 0.45 \\
        \bottomrule
    \end{tabular}
\end{table}


\paragraph{Direct-LLM diagnostic.}
An earlier zero-shot/direct-LLM run applied more than $20\,\mathrm{mm}$ under a
metric configuration that predates the control-cost definition used for the
matched ablations. Its scalar score is therefore not numerically comparable to
Table~\ref{tab:app-ablations} and is not used as quantitative evidence in the
main paper.


\section{Foundation-Model Robustness}
\label{app:model_robustness}
\label{app:model-robustness}

\subsection{Model Scale}
\begin{figure}[t]
    \centering
    \begin{subfigure}[b]{0.485\linewidth}
        \centering
        \includegraphics[width=\linewidth]{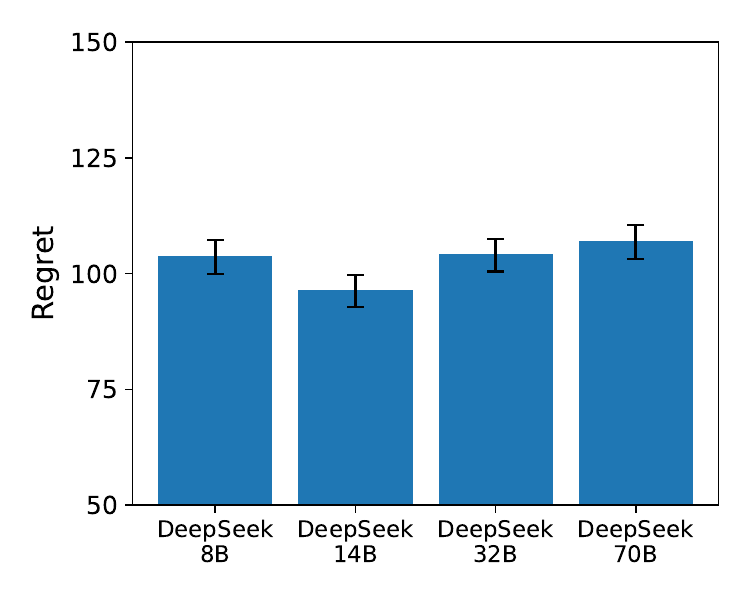}
        \caption{Control cost.}
    \end{subfigure}
    \hfill
    \begin{subfigure}[b]{0.485\linewidth}
        \centering
        \includegraphics[width=\linewidth]{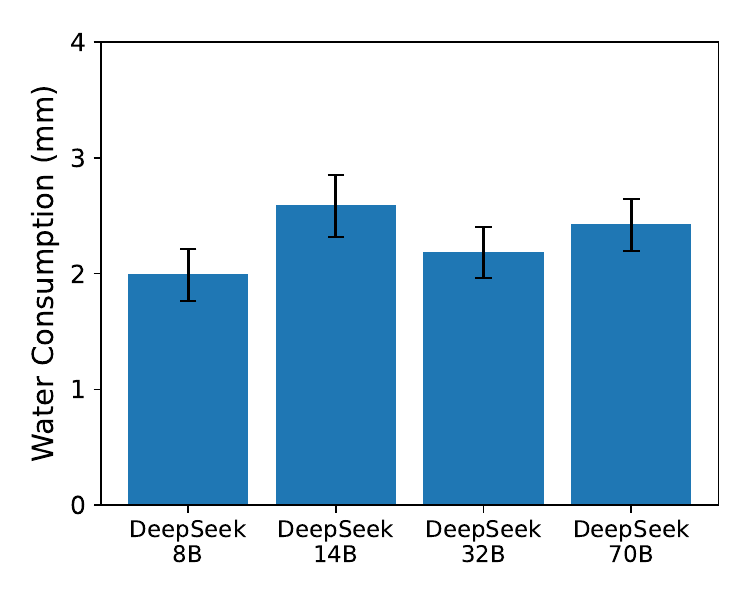}
        \caption{Irrigation applied.}
    \end{subfigure}
    \caption{DeepSeek scale sensitivity under an otherwise fixed \ours pipeline.}
    \label{fig:app-scale}
\end{figure}

DeepSeek-14B records $96.23\pm12$ control cost with
$2.585\,\mathrm{mm}$ irrigation in the reported sweep, whereas DeepSeek-8B
records $103.54\pm18$ with lower irrigation. The 32B and 70B variants remain in
a comparable range. We therefore report a non-monotonic relationship between
parameter count and control performance within these tested checkpoints.

\subsection{Model Family}
\begin{figure}[t]
    \centering
    \begin{subfigure}[b]{0.485\linewidth}
        \centering
        \includegraphics[width=\linewidth]{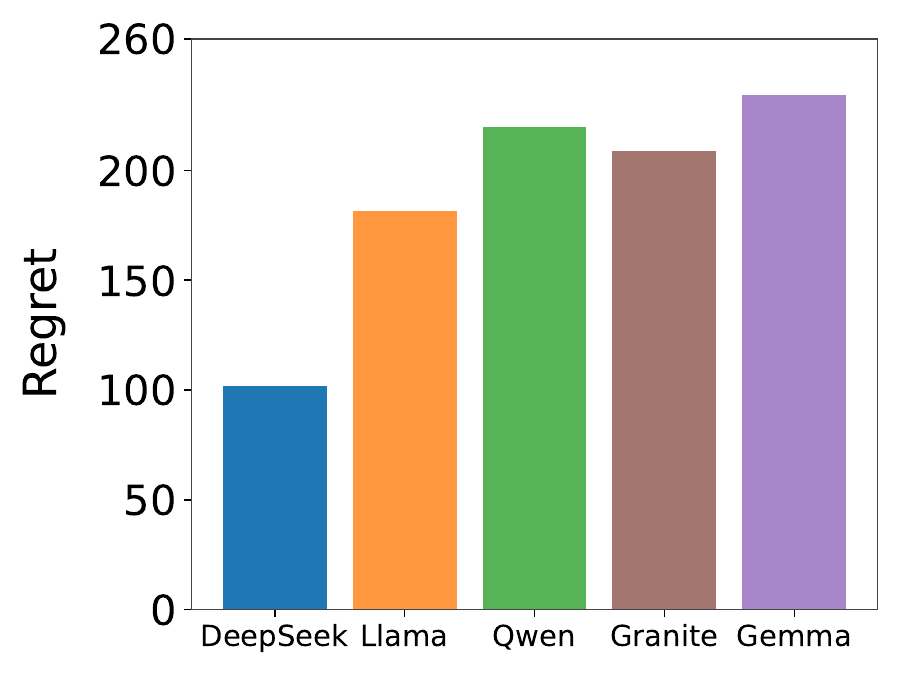}
        \caption{Control cost.}
    \end{subfigure}
    \hfill
    \begin{subfigure}[b]{0.485\linewidth}
        \centering
        \includegraphics[width=\linewidth]{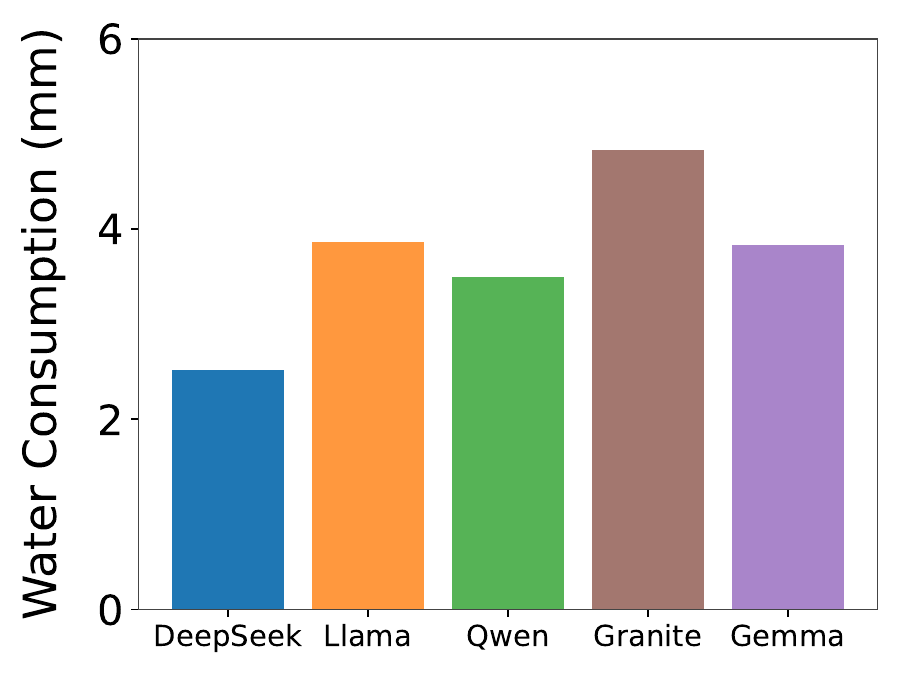}
        \caption{Irrigation applied.}
    \end{subfigure}
    \caption{Model-family sensitivity under a shared physical interface and evaluation sequence.}
    \label{fig:app-family}
\end{figure}

The tested families are DeepSeek, Qwen, Llama, Gemma, and Granite. DeepSeek has
the lowest explicitly reported value in this sweep
($101.92$ control cost, $4.26\,\mathrm{mm}$ irrigation), while the remaining
families remain in the same broad numerical range. These results are interpreted
as robustness/sensitivity within the tested set.


\section{Nine-Season Long-Horizon Analysis}
\label{app:long_horizon}
\label{app:nine-year}

\begin{figure}[t]
    \centering
    \begin{subfigure}[b]{0.485\linewidth}
        \centering
        \includegraphics[width=\linewidth]{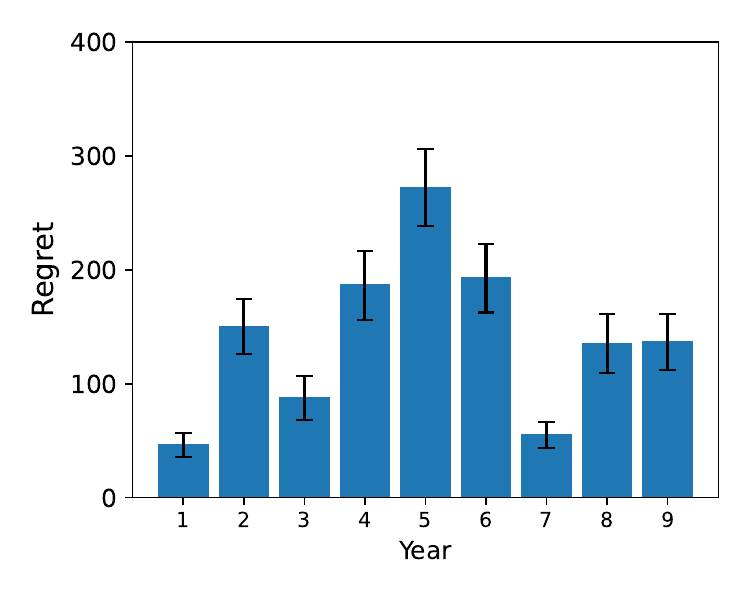}
        \caption{Annual control cost.}
    \end{subfigure}
    \hfill
    \begin{subfigure}[b]{0.485\linewidth}
        \centering
        \includegraphics[width=\linewidth]{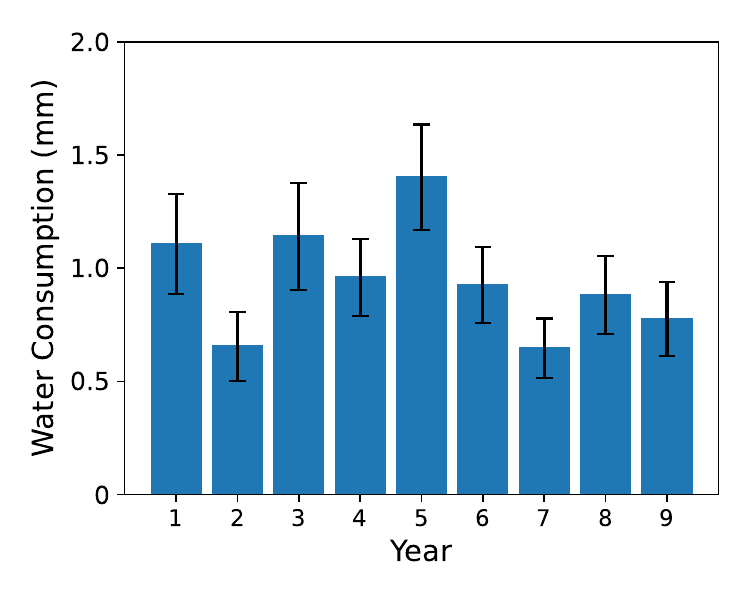}
        \caption{Annual irrigation applied.}
    \end{subfigure}
    \caption{Nine-season Colorado evaluation sequence.}
    \label{fig:app-nine-year}
\end{figure}

The largest explicitly reported annual cost is $272.15$ in Year~5. The
nine-season sequence is used as an operational-stability analysis: control cost
varies with seasonal forcing, but the sequence does not show sustained
divergence after the high-cost season. Individual years are not assigned a
causal weather label unless the corresponding weather evidence is reported
alongside the control result.

This analysis is intentionally separated from the adaptation ablation. The
sequence supports the claim that the complete system can continue operating
over heterogeneous seasons without persistent divergence; it does not by itself
demonstrate monotonic year-by-year learning. Evidence for the contribution of
persistent context comes from the matched ACE ablation in
Appendix~\ref{app:ablations}.


\section{Reproducibility and Provenance}
\label{app:reproducibility}

For each reported experiment, the evaluation pipeline records the
dataset/configuration identifier, software revision, language-model checkpoint,
random seed, and data hash. Non-LLM components are deterministic under a fixed
environment and configuration. Language-model calls use the frozen decoding
settings reported in Appendix~\ref{app:llm}.

The controller operates once per day while the simulator integrates the physical
state at hourly resolution. Reproduction should use the exact frozen
configuration associated with each reported result, including metric weights,
candidate-search span and step size, persistent-context thresholds and reset
policy, seed set, and learning-baseline hyperparameters.

The source code is intended for public release upon publication. Experiment
artifacts should be treated as the authoritative record for run-specific
configuration values.


\section{Discussion and Limitations}
\label{app:discussion}
\label{sec:discussion}

\paragraph{What the LLM contributes.}
The LLM in \ours is neither the physical model nor the final optimizer. It maps
structured state, forecast, crop-stage information, and persistent guidance to a
context-conditioned proposal and can revise that proposal after numerical
feedback. Physical rollout evaluates consequences, deterministic selection ranks
bounded candidates, and runtime assurance retains final execution authority.
The paper therefore studies how adaptive semantic reasoning can be composed with
trusted numerical control; it does not establish that an LLM is necessary for
every irrigation problem or that it is superior to the best deterministic
optimizer.

\paragraph{Bounded self-improvement and stopping conditions.}
\ours implements scaffold-level, experience-driven adaptation. 
Persistent principles can change future proposals,
but updates are evidence-triggered and cannot modify model weights, the physical
transition model, the evaluator, or the feasible-action set. Fast-loop repair
terminates when physical verification no longer changes the proposal or the fixed
round budget is reached. Slow-loop adaptation is a no-op when no recurring
pattern satisfies the configured update trigger. These semantics prevent
open-ended reflection from becoming part of actuation.

We do not claim that persistent memory converges, improves monotonically, or is
immune to forgetting. \ours should therefore be viewed as a bounded step toward
persistent or lifelong physical agency.

\paragraph{Implication for embodied physical agents.}
Agriculture is a persistent embodied-control problem: actions modify a physical
state, feedback is delayed, environmental forcing changes over time, and
experience accumulates across repeated interaction\cite{Adonis25Liu,hydra24liu, Proteus25Liu, liu2025hydrabench, liu26aura, gan2023poster}.. The transferable design
lesson is a separation of authority across timescales: semantic reasoning may
propose and adapt; numerical models predict consequences; deterministic logic
constrains execution; and persistent context can repair future reasoning without
rewriting the trusted physical substrate. Whether the same decomposition
transfers to other long-horizon physical domains remains an empirical question
for future work.


\section{AI Use Disclosure}
\label{app:ai_use}
\label{sec:ai_use}
We used generative AI tools, including OpenAI ChatGPT, OpenAI Codex, and Google Gemini, during the development and preparation of this work.

These tools were used for four main purposes. First, they assisted with writing and editing, including improving the clarity, readability, organization, and presentation of author-provided technical content. Second, they supported literature retrieval and discovery by helping identify potentially relevant references and organize related-work discussions. Third, they assisted with research ideation and execution, including organizing research logic, translating author-defined objectives and conceptual designs into implementations, providing feedback on experimental design, and supporting code development and debugging. Fourth, they were used to help draft and revise portions of the manuscript based on research objectives, outlines, methods, experimental results, and technical content specified by the authors.

Generative AI was not used to fabricate experimental observations, numerical results, or evaluation outcomes. All empirical data reported in this work originate from the real-world measurements or research datasets described in the paper, and all reported quantitative results were obtained by executing the implemented experimental pipeline.

All AI-assisted code, experimental configurations, numerical results, citations, technical statements, and scientific claims were reviewed and verified by the authors. Suggested references were checked against their original sources, and AI-generated or AI-assisted code used in the reported experiments was inspected and tested before use.

Generative AI tools served only as research and writing assistants. The authors determined the research questions, methodological choices, experimental protocols, analyses, interpretations, claims, and conclusions. The authors take full responsibility for the final content of this work, including all text, code, results, citations, claims, and other artifacts produced with the assistance of generative AI.

\end{document}